\documentclass{bmvc2k}

\usepackage{wrapfig,lipsum,booktabs}
\usepackage{subcaption}
\usepackage{enumitem}
\title{Geometry-Aware Multi-Task Learning for\\ Binaural Audio Generation from Video}

\addauthor{Rishabh Garg\textsuperscript{1}}{rishabh@cs.utexas.edu}{1}
\addauthor{Ruohan Gao\textsuperscript{2}}{rhgao@cs.stanford.edu}{2}
\addauthor{Kristen Grauman\textsuperscript{1,3}}{grauman@cs.utexas.edu}{3}

\addinstitution{
 \textsuperscript{1}The University of Texas at Austin
}
\addinstitution{
 \textsuperscript{2}Stanford University
}
\addinstitution{
 \textsuperscript{3}Facebook AI Research
}

\usepackage[font=small,labelfont=bf]{caption}

\runninghead{Garg et al}{Geometry-Aware Multi-Task Binaural Generation from Video}

\def\etal{\emph{et al}\bmvaOneDot}
\begin{document}

\maketitle

%===========================================================
\begin{abstract}
Binaural audio provides human listeners with an immersive spatial sound experience, but most existing videos lack binaural audio recordings.  We propose an audio spatialization method that draws on visual information in videos to convert their monaural (single-channel) audio to binaural audio.  Whereas existing approaches leverage visual features extracted directly from video frames, our approach explicitly disentangles the geometric cues present in the visual stream to guide the learning process.  In particular, we develop a multi-task framework that learns geometry-aware features for binaural audio generation by accounting for the  underlying room impulse response, the visual stream's coherence with the sound source(s) positions, and the consistency in geometry of the sounding objects over time.  Furthermore, we introduce a new large video dataset with realistic binaural audio simulated for real-world scanned environments.  On two datasets, we demonstrate the efficacy of our method, which achieves state-of-the-art results.
\end{abstract}

%===========================================================
%-------------------------------------------------------------------------
\section{Introduction}
\label{sec:intro}

Both sight and sound are key drivers of the human perceptual experience, and both convey essential spatial information.  For example, a car driving past us is audible---and spatially trackable---even before it crosses our field of view; a bird singing high in the trees helps us spot it with binoculars; a chamber music quartet performance sounds spatially rich, with the instruments' layout on stage affecting our listening experience.  

Spatial hearing is possible thanks to the \emph{binaural} audio received by our two ears.  The Interaural Level Difference (ILD) and the Interaural Time Difference (ITD) between the sounds reaching each ear, as well as the shape of the outer ears themselves, all provide spatial effects~\cite{rayleigh1875our}.  Meanwhile, the reflections and reverberations of sound in the environment are a function of the room acoustics---the geometry of the room, its major surfaces, and their materials.  For example, we perceive  the same audio differently in a long corridor versus a large room, or a room with heavy carpet versus a smooth marble floor.

\begin{figure*}[t]
\centering
\includegraphics[width=1\textwidth]{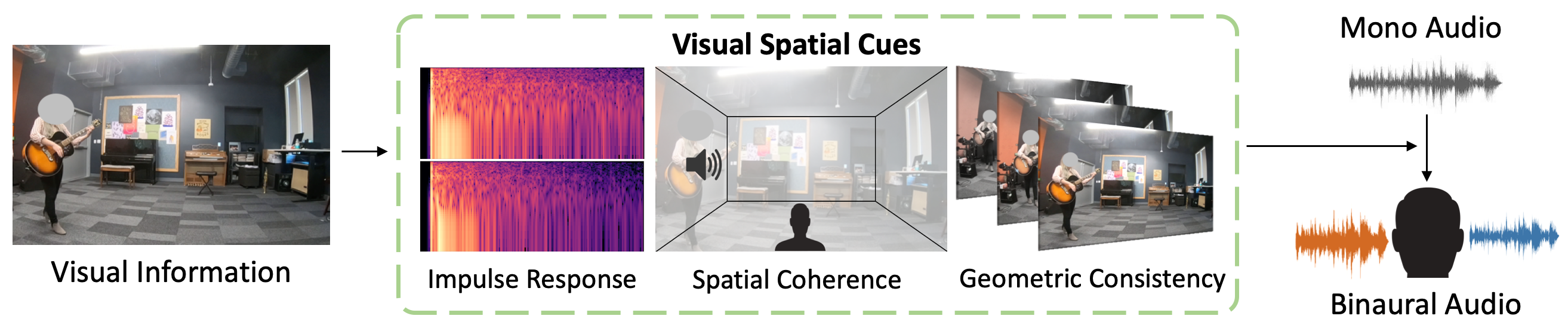}
\vspace*{0.1in}
\caption{To generate accurate binaural audio from monaural audio, the visuals provide significant cues that can be learnt jointly with audio prediction.  Our approach learns to extract spatial information (e.g., the guitar player is on the left), geometric consistency of the position of the sound sources over time, and cues from the inferred binaural impulse response from the surrounding room. 
}
\label{fig:teaser}
\end{figure*}

Videos or other media with binaural audio imitate that rich audio experience for a user, making the media feel more real and immersive. This immersion is important for  virtual reality and augmented reality applications, where the user should feel transported to another place and perceive it as such. However, collecting binaural audio data is a challenge. Presently, spatial audio is collected with an array of microphones or specialized dummy rig that imitates the human ears and head.  The collection process is therefore less accessible and more costly compared to standard single-channel monaural audio captured with ease from today's ubiquitous mobile devices.

Recent work explores how monaural audio can be upgraded to binaural audio by leveraging the \emph{visual} stream in videos~\cite{morgadoNIPS18, gao2019visualsound, zhou2020sep}. The premise is that the visual context provides hints for how to spatialize the sound due to the visible sounding objects and room geometry.  While inspiring, existing models are nonetheless limited to extracting generic visual cues that only implicitly infer spatial characteristics.

Our idea is to explicitly model the spatial phenomena in video that influence the associated binaural sound. Going beyond generic visual features, our approach guides binauralization with those geometric cues from the object and environment that dictate how a listener receives the sound in the real world. In particular, we introduce a multi-task learning framework that accounts for three key factors (Fig.~\ref{fig:teaser}). First, we require the visual features to be predictive of the \emph{room impulse response} (RIR), which is
the transfer function between the sound sources, 3D environment, and camera/microphone position.  
Second, we require the visual features to be \textit{spatially coherent} with the sound, i.e., they can understand the difference when audio is aligned with the visuals and when it is not. 
Third, we enforce the \emph{geometric consistency} of objects over time in the video.  Whereas existing methods treat audio and visual frame pairs as independent samples, our approach represents the spatio-temporal smoothness of objects in video, which generally do not have dramatic instantaneous changes in their layout.

The main contributions of this work are as follows.
Firstly, we propose a novel multi-task approach to convert a video's monaural sound to binaural sound  by learning audio-visual representations that leverage geometric characteristics of the environment and the spatial and temporal cues from videos. Second, to facilitate binauralization research, we create SimBinaural, a large-scale dataset of simulated videos with binaural sound in photo-realistic 3D indoor scene environments. This new dataset facilitates both learning and quantitative evaluation, allows us to explore the impact of particular parameters in a controlled manner, and even benefits learning in real videos. Finally, we show the efficacy of our method via extensive experiments in generating realistic binaural audio, achieving state-of-the-art results.
%===========================================================
\vspace*{-0.1in}
\section{Related Work}

\vspace*{-0.05in}
\paragraph{Visually-Guided Audio Spatialization}

Recent work uses video frames to provide a form of self-supervision to implicitly infer the relative positions of sound-making objects. They formulate the problem as an upmixing task from mono to binaural using the visual information.  Morgado \etal \cite{morgadoNIPS18} use 360 videos from YouTube to predict first order ambisonic sound useful for 360 viewing, while Lu \etal \cite{lu2019self} use a self-supervised audio spatialization network using visual frames and optical flow. Whereas \cite{lu2019self} uses correspondence to learn audio synthesizer ratio masks, which does not necessitate understanding of sound making objects, we enforce understanding of the sound location via spatial coherence in the visual features. For speech synthesis, using the ground truth position and orientation of the source and receiver instead of a video is also explored~\cite{richard2021binaural}.

More closely related to our problem, the 2.5D visual sound approach by Gao and Grauman generates binaural audio from video~\cite{gao2019visualsound}. Building on those ideas, Zhou \etal \cite{zhou2020sep} propose an associative pyramid network (APNet) architecture to fuse the modalities and jointly train on audio spatialization and source separation task. Concurrent to our work, Xu \etal \cite{xu2021visually} propose to generate binaural audio for training from mono audio by using spherical harmonics. In contrast to these methods, we explore a novel framework for learning geometric representations, and we introduce  a large-scale photo-realistic video dataset with acoustically accurate binaural information (which will be shared publicly). We outperform the existing methods and show that the new dataset can be used to augment performance.

\vspace*{-0.2in}
\paragraph{Audio and 3D Spaces}
Recent work exploits the complementary nature of audio and the characteristics of the environment in which it is heard or recorded. Prior methods estimate the acoustic properties of materials \cite{schissler2017acoustic}, estimate reverberation time and equalization of the room using an actual 3D model of a room \cite{tang2020scene}, and learn audio-visual correspondence from video \cite{yang2020telling}. Chen \etal~\cite{chen2020soundspaces} introduce the SoundSpaces audio platform to perform audio-visual navigation in scanned 3D environments, using binaural audio to guide policy learning. Ongoing work continues to explore audio-visual navigation models for embodied agents~\cite{gan2020look,dean,chen2020learning, chen2020semantic,sagnik-move2hear}. Other work predicts depth maps using spatial audio \cite{christensen2020batvision} and learns representations  via interaction using echoes recorded in indoor 3D simulated environments~\cite{gao2020visualechoes}. In contrast to all of the above, we are interested in a different problem of generating accurate spatial binaural sound from videos. We do not use it for navigation nor to explicitly estimate information about the environment.  Rather, the output of our model is spatial sound to provide a human listener with an immersive audio-visual experience. 

\vspace*{-0.2in}
\paragraph{Audio-Visual Learning} 
Audio-visual learning has a long history, and has enjoyed a resurgence in the vision community in recent years.  Cross-modal learning is explored to understand the natural synchronisation between visuals and the audio \cite{arandjelovic2017look, aytar2016soundnet, owens2016ambient}. Audio-visual data is leveraged for audio-visual speech recognition \cite{hu2016temporal, chung2017lip, zhou2019talking, yu2020audio}, audio-visual event localization \cite{tian2018audio, tian2020unified, wu2019dual}, sound source localization \cite{arandjelovic2017objects, zhao2018sound, Senocak_2018_CVPR, tian2018audio, hu2020discriminative, rouditchenko2019self}, self-supervised representation learning \cite{owens2016ambient, owens2018audio, Korbar2018cotraining, gao2020visualechoes, morgado2020learning}, generating sounds from video \cite{owens2016visually, zhou2017visual, gan2020foley, chen2020generating}, and audio-visual source separation for speech \cite{ephrat2018looking, owens2018audio, afouras2018conversation, gabbay2017visual, afouras2019my, chung2020facefilter}, music \cite{zhao2018sound, zhao2019som, xu2019recursive, gan2020music, gao2019coseparation}, and objects \cite{gao2018objectSounds, gao2019coseparation, tzinis2020into}. In contrast to all these methods, we perform a different task: to produce binaural two-channel audio from a monaural audio clip using a video's visual stream.

%===========================================================
 \vspace*{-0.3in}
\section{Approach}

Our goal is to generate binaural audio from videos with monaural audio. In this section, we first formally describe the problem  (Section~\ref{sec:task}).  Then we introduce our proposed multi-task setting (Section~\ref{sec:multitask}). Next we describe the training and inference method (Section~\ref{sec:train_inf}), and finally we describe the proposed SimBinaural dataset (Section~\ref{sec:simBinaural}).

\vspace*{-0.05in}
\subsection{Problem Formulation}
\label{sec:task}
Our objective is to map the monaural sound from a given video to spatial binaural audio. The input video may have one or more sound sources, and neither their positions in the 3D scene nor their positions in the 2D video frame are given.

For a video $\mathcal{V}$ with frames $\{v^1...v^T\}$ and  monaural audio $a_M^t$, we aim to predict a two channel binaural audio output $\{a_L^t, a_R^t\}$. Whereas a single-channel audio $a^t_M$ lacks spatial characteristics, two-channel binaural audio $\{a^t_L, a^t_R\}$ conveys two distinct waveforms to the left and right ears separately and hence provides spatial effects to the listener.  By coupling the monaural audio with the visual stream, we aim to leverage the spatial cues from the pixels to infer how to spatialize the sound. We first transfer the input audio waveforms into the time-frequency domain using the Short-Time Fourier Transformation (STFT). We aim to predict the binaural audio spectrograms $\{ \mathcal{A}^t_L, \mathcal{A}^t_R \}$ from the input mono spectrogram $\mathcal{A}^t_M$, where $\mathcal{A}^t_X = \text{STFT}(a^t_X)$, conditioned on visual features $v_f^t$ from the video frames at time $t$. 

\vspace*{-0.1in}
\subsection{Geometry-Aware Multi-Task Binauralization Network}
\label{sec:multitask}

Our approach has four main components: the \emph{backbone} for converting mono audio to binaural by injecting the visual information, the \emph{spatial coherence} module that learns the relative alignment of the spatial sound and frame,  an \emph{RIR prediction} module that requires the room impulse response to be predictable from the video frames, and the \emph{geometric consistency} module that enforces consistency of objects over time.

\begin{figure*}[t]
    \centering
    \includegraphics[scale=0.44]{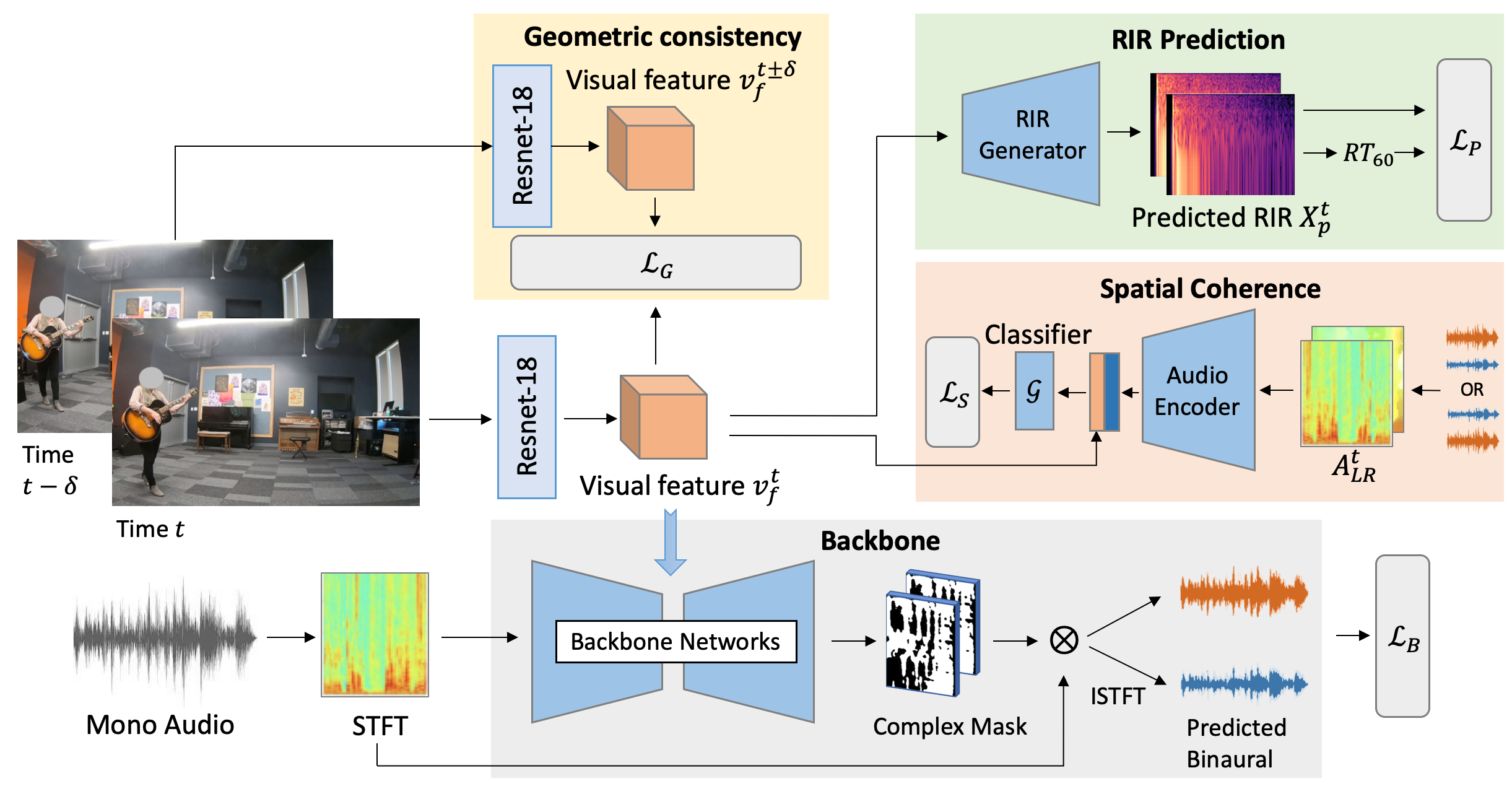}
    \vspace*{0.1in}
    \caption{Proposed network. The network takes the visual frames and monaural audio as input. The ResNet-18 visual features $v_f^t$ are trained in a multi-task setting. The features $v_f^t$ are used to directly predict the RIR via a decoder (top right). Audio features from binaural audio, which might have flipped channels, are combined with $v_f^t$ and used to train a spatial coherence classifier $\mathcal{G}$ (middle right). Two temporally adjacent frames are also used to ensure geometric consistency (top center). The features $v_f^t$ are jointly trained with the backbone network (bottom) to predict the final binaural audio output.
    }
    \label{fig:main}
\end{figure*}

\paragraph{Backbone Loss}
First, we define the backbone loss within our multi-task framework (Fig.~\ref{fig:main}, bottom). 
This backbone network is used to transform the input monaural spectrogram $\mathcal{A}^t_M$ to binaural ones. During training, the mono audio is obtained by averaging the two channels $a_M^t = (a_L^t + a_R^t)/2$ and hence the spatial information is lost. Rather than directly predict the two channels of binaural output, we predict the \emph{difference} of the two channels, following~\cite{gao2019visualsound}. This better captures the subtle distinction of the channels and avoids collapse to the easy case of predicting the same output for both channels. We predict a complex mask $M_D^t$, which, multiplied with the original audio spectrogram $\mathcal{A}^t_M$, gives the predicted difference spectrogram $\mathcal{A}^t_{D(pred)} = M_D^t \cdot \mathcal{A}^t_M$. The true difference spectrogram of the training input $\mathcal{A}_D^t$ is the STFT of $a_L^t - a_R^t$. We minimize the distance between these two spectrograms: $\|\mathcal{A}_D^t - \mathcal{A}_{D(pred)}^t \|_2^2$.  We also predict the two channels via two complex masks $M_L^t$ and $M_R^t$, one for each channel, to obtain the predicted channel spectrograms $\mathcal{A}_{L(pred)}^t$ and $\mathcal{A}_{R(pred)}^t$ like above. 
This gives us the overall backbone objective: 
\vspace*{-0.1in}
\begin{equation}
    \mathcal{L}_B = \| \mathcal{A}_D^t - \mathcal{A}_{D(pred)}^t \|_2^2 +  \Big\{ \| \mathcal{A}_L^t - \mathcal{A}_{L(pred)}^t \|_2^2  + \| \mathcal{A}_R^t - \mathcal{A}_{R(pred)}^t \|_2^2 \Big\}.
\label{eq:base}
\vspace*{-0.05in}
\end{equation}

\paragraph{Spatial Coherence} 
We encourage the visual features to have geometric understanding of the relative positions of the sound source and receiver via an audio-visual feature alignment prediction term. This loss requires the predicted audio to correctly capture which channel is left and right with respect to the visual information. This is crucial to achieve the proper spatial effect while watching videos, as the audio needs to match the observed visuals' layout.

In particular, we incorporate a classifier to identify whether the visual input is aligned with the audio. The classifier $\mathcal{G}$ combines the binaural audio $\mathcal{A}_{LR} = \{\mathcal{A}_L^t, \mathcal{A}_R^t\}$ and the visual features $v_f^t$ to classify if the audio and visuals agree.   In this way, the visual features are forced to reason about the relative positions of the sound sources and learn to find the cues in the visual frames which dictate the direction of sound heard. During training, the original ground truth samples are aligned, and we create misaligned samples by flipping the two channels in the ground truth audio to get $\mathcal{A}_{LR} = \{\mathcal{A}_R^t, \mathcal{A}_L^t\}$. We calculate the binary cross entropy (BCE) loss for the classifier's prediction of whether the audio is flipped or not, $c = \mathcal{G}(\mathcal{A}_{LR}, v_f^t)$, and the  indicator $\hat{c}$ denoting if the audio is flipped, yielding the spatial coherence loss:
\vspace*{-0.05in}
\begin{equation}
    \mathcal{L}_G = \text{BCE}(\mathcal{G}(\mathcal{A}_{LR}, v_f^t), \; \hat{c}).
\label{eq:geom}
\vspace*{-0.1in}
\end{equation}

\paragraph{Room Impulse Response and Reverberation Time Prediction} 
The third  component of our multi-task model trains the visual features to be predictive of the room impulse response (RIR). An impulse response gives a concise acoustic description of the environment, consisting of the initial direct sound, the early reflections from the surfaces of the room, and a reverberant tail from the subsequent higher order reflections between the source and receiver. The visual frames convey information like the layout of the room and the sound source with respect to the receiver, which in part form the basis of the RIR. Since we want our audio-visual feature to be a latent representation of the geometry of the room and the source-receiver position pair, we introduce an auxiliary task to predict the room IR directly from the visual frames via a generator on the visual features.  Furthermore, we require the features to be predictive of the \emph{reverberation time} $RT_{60}$ metric, which is the time it takes the energy of the impulse to decay 60dB, and can be calculated from the energy decay curve of the IR \cite{schroeder1965new}.  The $RT_{60}$ is commonly used to characterize the sound properties of a room; we employ it as a low-dimensional target here to guide feature learning alongside the high-dimensional RIR spectrogram prediction.

We convert the ground truth binaural impulse response signal $\{r_L ,r_R\}$ to the frequency domain using the STFT and obtain magnitude spectrograms $\mathcal{X}$ for each channel. The IR prediction network consists of a generator which performs upconvolutions on the visual features $v_f^t$  to obtain a predicted magnitude spectrogram $\mathcal{X}_{(pred)}^t$. We minimize the euclidean distance between the predicted RIR $\mathcal{X}^t_{(pred)}$, and the ground truth $\mathcal{X}^t_{gt}$. Additionally, we obtain the RIR waveform from the predicted spectrogram $\mathcal{X}^t_{(pred)}$ via the Griffin-Lim algorithm \cite{griffin1984signal, perraudin2013fast} and compute the $RT_{60(pred)}$. We minimize the L1 distance between the predicted  $RT_{60(pred)}$ and the ground truth $RT_{60(gt)}$. Thus, the overall RIR prediction loss is:
\vspace*{-0.1in}
\begin{equation}
    \mathcal{L}_P = \| \mathcal{X}^t_{(pred)} - \mathcal{X}^t_{gt} \|_2^2 + | RT_{60(pred)} - RT_{60(gt)} |.
\label{eq:ir}
\end{equation}

\paragraph{Geometric Consistency}
Since the videos are continuous samples over time rather than individual frames, our fourth and final loss regularizes the visual features by requiring them to have spatio-temporal geometric consistency.  The position of the source(s) of sound and the position of the camera---as well as the physical environment where the video is recorded---do not typically change instantaneously in videos. Therefore, there is a natural coherence between the sound in a video observed at two points that are temporally close. Since visual features are used to condition our binaural prediction, we encourage our visual features to learn a latent representation that is coherent across short intervals of time. Specifically, the visual features $v_f^t$ and $v_f^{t\pm\delta}$ should be relatively similar to each other to produce audio with fairly similar spatial effects. Specifically, the geometric consistency loss is: 
\vspace*{-0.05in}
\begin{equation}
    \mathcal{L}_S = \text{max}( \| v_f^t - v_f^{t\pm\delta} \| - \alpha, 0),
\label{eq:spatial}
\vspace*{-0.05in}
\end{equation}
where $\alpha$ is the margin allowed between two visual features. We select a random frame $\pm 1$ second from $t$, so $-1 \leq \delta \leq 1$. This ensures that similar placements of the camera with respect to the audio source should be represented with similar features, while the margin allows room for dissimilarity for the changes due to time. Since the underlying visual features are  regularized to be similar, the predicted audio conditioned on these visual features is also encouraged to be temporally consistent.

\subsection{Training and Inference}
\label{sec:train_inf}
During training, the mono audio is obtained by taking the mean of the two channels of the ground truth audio $a_m^t = (a_L^t + a_R^t)/2$. The visual features $v_f^t$ are reduced in dimension, tiled, and concatenated with the output of the audio encoder to fuse the information from the audio and visual streams. The overall multi-task loss is a combination of the losses (Equations \ref{eq:base}-\ref{eq:spatial}) described earlier:
\vspace*{-0.05in}
\begin{equation}
    \mathcal{L} = \lambda_B \mathcal{L}_B + \lambda_S \mathcal{L}_S + \lambda_G \mathcal{L}_G + \lambda_P\mathcal{L}_P,
\vspace*{-0.05in}
\label{eq:full}
\end{equation}
where $\lambda_B$, $\lambda_S$, $\lambda_G$ and $\lambda_P$ are the scalar weights used to determine the effect of each loss during training, set using validation data.  To generate audio at test time, we only require the mono audio and visual frames. The predicted spectrograms are used to obtain the predicted difference signal $a^t_{D(pred)}$ and two-channel audio $\{a^t_L, a^t_R\}$ via an inverse Short-Time Fourier Transformation (ISTFT) operation.

\vspace*{-0.1in}
\subsection{SimBinaural Dataset}
\label{sec:simBinaural}
 
\begin{figure*}[t]
\centering
\includegraphics[width=\textwidth]{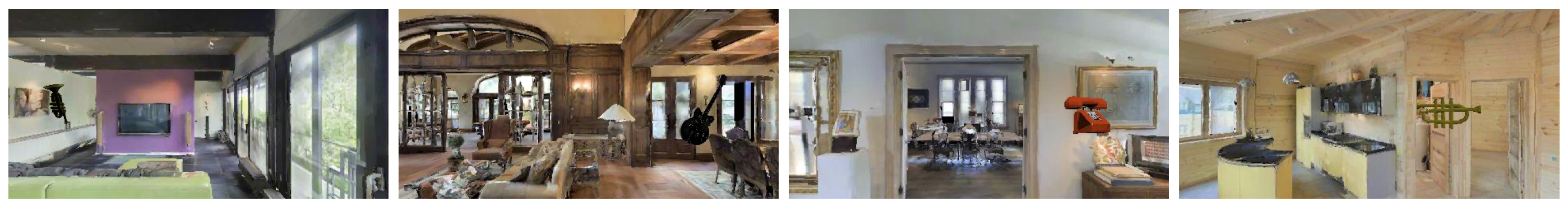}
\vspace{-0.1in}
\caption{Example frames from the SimBinaural dataset.}
\label{fig:examples}
\vspace{-0.1in}
\end{figure*}

We experiment with both real world video (FAIR-Play~\cite{gao2019visualsound}) and video from scanned environments with high quality simulated audio.  For the latter, to facilitate large-scale experimentation---and to augment learning from real videos---we create a new dataset called SimBinaural of simulated videos in photo-realistic 3D indoor scene environments.\footnote{The SimBinaural dataset was constructed at, and will be released by, The University of Texas at Austin.}  The generated videos, totalling over 100 hours, resemble real-world audio recordings and are sampled from 1,020 rooms in 80 distinct environments; each environment is a multi-room home. Using the publicly available SoundSpaces\footnote{SoundSpaces~\cite{chen2020soundspaces} provides room impulse responses at a spatial resolution of 1 meter. These state-of-the-art RIRs capture how sound from each source propagates and interacts with the surrounding geometry and materials, modeling all of the major real-world features of the RIR: direct sounds, early specular/diffuse reflections, reverberations, binaural spatialization, and frequency dependent effects from materials and air absorption.}  audio simulations~\cite{chen2020soundspaces} together with the Habitat simulator~\cite{savva2019habitat}, we create realistic videos with binaural sounds for publicly available 3D environments in Matterport3D \cite{chang2017matterport3d}. See Fig.~\ref{fig:examples} and Supp.~video.  Our resulting dataset is much larger and more diverse than the widely used FAIR-Play dataset \cite{gao2019visualsound} which is real video but is limited to 5 hours of recordings in one room (Table \ref{tab:data_comparison}).

To construct the dataset, we insert diverse 3D models from \texttt{poly.google.com} of various instruments like guitar, violin, flute etc. and other sound-making objects like phones and clocks into the scene. To generate realistic binaural sound in the environment as if it is coming from the source location and heard at the camera position, we convolve the appropriate SoundSpaces~\cite{chen2020soundspaces} room impulse response with an anechoic audio waveform (e.g., a guitar playing for an inserted guitar 3D object). Using this setup, we capture videos with simulated binaural sound. The virtual camera and attached microphones are moved along trajectories such that the object remains in view, leading to diversity in views of the object and locations within each video clip. Please see Supp.~for details.

\begin{table}[t]\small
\centering
    \begin{tabular}{c|cccc}
    Dataset        & \#Videos & Length (hrs) & \#Rooms & RIR \\ \hline
    FAIR-Play~\cite{gao2019visualsound} & 1,871   & 5.2  & 1 & No \\
    SimBinaural & 21,737 & 116.1        & 1,020 & Yes
    \end{tabular}
\vspace*{0.1in}
\caption{A comparison of the data in FAIR-Play and the large scale data we generated.}
\label{tab:data_comparison}
\vspace{-0.1in}
\end{table}
%===========================================================
\vspace*{-0.1in}
\section{Experiments}\label{sec:exp}
\vspace*{-0.05in}

We validate our approach on both FAIR-Play~\cite{gao2019visualsound} (an existing real video benchmark) and our new SimBinaural dataset. We compare  to the following baselines:
\vspace*{-0.075in}
\begin{itemize}
\item \textbf{Flipped-Visual:} We flip the visual frame horizontally to provide incorrect visual information while testing. The other settings are the same as our method.\vspace*{-0.075in}
\item \textbf{Audio Only:} We provide only monaural audio as input, with no visual frames, to verify if the visual information is essential to learning.\vspace*{-0.075in}
\item \textbf{Mono-Mono:} Both channels have the same input monaural audio repeated as the two-channel output to verify if we are actually distinguishing between the channels.\vspace*{-0.075in}
\item \textbf{Mono2Binaural~\cite{gao2019visualsound}}: A state-of-the-art 2.5D visual sound model for this task.  We use the authors' code to evaluate in the settings as ours.\vspace*{-0.075in}
\item \textbf{APNet~\cite{zhou2020sep}}: A state-of-the-art model that handles both binauralization and audio source separation. We use the APNet network from their method and train only on binaural data (rather than stereo audio).  We use the authors' code.\vspace*{-0.075in}
\item \textbf{PseudoBinaural \cite{xu2021visually}}: A state-of-the-art model that uses additional data to augment training. We use the authors' public pre-trained model.\vspace*{-0.075in}
\end{itemize}

We evaluate two standard metrics, following~\cite{gao2019visualsound, morgadoNIPS18, zhou2020sep}: 1) \textbf{STFT Distance}, the euclidean distance between the predicted and ground truth STFT spectrograms, which directly measures how accurate is our produced spectrogram, 2) \textbf{Envelope Distance} (ENV) which measures the euclidean distance between the envelopes of the predicted raw audio signal and the ground truth and can further capture the perceptual similarity.

\vspace*{-0.1in}
\paragraph{Implementation details}
All networks are written in PyTorch \cite{NEURIPS2019_9015}. The backbone network is based upon the networks used for 2.5D visual sound~\cite{gao2019visualsound} and APNet~\cite{zhou2020sep}. The audio network consists of a U-Net \cite{ronneberger2015u} type architecture while the RIR generator is adapted from GANSynth \cite{engel2019gansynth}. To preprocess both datasets, we follow the standard steps from~\cite{gao2019visualsound}. We resample all the audio to 16kHz and for training the backbone, we use 0.63s clips of the 10s audio and the corresponding frame. Frames are extracted at 10fps. The visual frames are randomly cropped to 448 $\times$ 224. For testing, we use a sliding window of 0.1s to compute the binaural audio for all methods. Please see Supp.~for more details.

\begin{table}[t]\small
\centering
\begin{tabular}{ccccccc}
\multicolumn{1}{l}{} & \multicolumn{2}{c}{FAIR-Play} & \multicolumn{4}{c}{SimBinaural} \\ 
\cline{4-7} & \multicolumn{1}{l}{} & \multicolumn{1}{l}{} & \multicolumn{2}{c}{Scene-Split} & \multicolumn{2}{c}{Position-Split} \\
                                       & STFT      & ENV               & STFT   & ENV    & STFT  & ENV   \\ \hline
Mono-Mono                              & 1.215     & 0.157             & 1.356  & 0.163  & 1.348 & 0.168 \\
Audio-Only                             & 1.102     & 0.145             & 0.973  & 0.135  & 0.932 & 0.130 \\
Flipped-Visual                         & 1.134     & 0.152             & 1.082  & 0.142  & 1.075 & 0.141 \\
Mono2Binaural~\cite{gao2019visualsound}& 0.927     & 0.142             & 0.874  & 0.129  & 0.805 & 0.124 \\
APNet \cite{zhou2020sep}               & 0.904     & 0.138             & 0.857  & 0.127  & 0.773 & 0.122 \\ \hline
Backbone+IR Pred                       & n/a       & n/a               & 0.801  & 0.124  & 0.713 & 0.117 \\
Backbone+Spatial                       & 0.873     & \textbf{0.134}    & 0.837  & 0.126  & 0.756 & 0.120 \\
Backbone+Geom                          & 0.874     & 0.135             & 0.828  & 0.125  & 0.731 & 0.118 \\
\textbf{Our Full Model} & \textbf{0.869}     & \textbf{0.134}       & \textbf{0.795} & \textbf{0.123} & \textbf{0.691}   & \textbf{0.116}  \\ \hline
\end{tabular}
\vspace*{0.1in}
\caption{Binaural audio prediction errors on the FAIR-Play and SimBinaural datasets. For both  metrics, lower is better.}
\label{tab:FairPlay-Metric}
\label{tab:SimBinaural-Metric}
\vspace{-0.1in}
\end{table}

\vspace*{-0.15in}
\paragraph{SimBinaural results} 
We evaluate on two data splits: 1) \textbf{\textit{Scene-Split}}, where the train and test set have disjoint scenes from Matterport3D \cite{chang2017matterport3d} and hence the room of the videos at test time has not been seen before; and 2) \textbf{\textit{Position-Split}}, where the splits may share the same Matterport3D scene/room but the exact configuration of the source object and receiver position is not seen before.  

Table~\ref{tab:SimBinaural-Metric} (right) shows the results.   The table also ablates the parts of our model. Our model outperforms all the baselines, including the two state-of-the-art prior methods. In addition, Table~\ref{tab:SimBinaural-Metric} confirms that Scene-Split is a fundamentally harder task. This is because we must predict the sound, as well as other characteristics like the IR, from visuals distinct from those we have observed before. This forces the model to generalize its encoding to generic visual properties (wall orientations, major furniture, etc.) that have intra-class variations and geometry changes compared to the training scenes.

The ablations shed light on the impact of each of the proposed losses in our multi-task framework. The full model uses all the losses as in Eqn~\ref{eq:full}. This outperforms other methods significantly on both splits. It also outperforms using each of the losses individually, which demonstrates the losses can combine to jointly learn better visual features for generating spatial audio. 

\vspace{-0.15in}
\paragraph{FAIR-Play results} 
\begin{wraptable}{r}{0.5\textwidth}
\centering
\begin{tabular}{ccc}\small
Method  & STFT   & ENV    \\ \hline
APNet~\cite{zhou2020sep}                                        & 1.291  & 0.162  \\
PseudoBinaural \cite{xu2021visually} & 1.268  & 0.161  \\
Ours                                         & 1.234  & 0.160  \\
\textbf{Ours+SimBinaural}                    & \textbf{1.175} & \textbf{0.154}
\end{tabular}
\vspace*{0.1in}
\caption{
Results on FAIR-Play when additional data is used for training.}
\vspace{-0.1in}
\label{tab:joint}
\end{wraptable}
Table \ref{tab:FairPlay-Metric} (left) shows the results on the real video benchmark FAIR-Play using the standard split. Here, we omit the IR prediction network for our method, since FAIR-Play lacks  ground truth impulse responses (which we need for training). The Backbone+Spatial and Backbone+Geom are the same as above. Both variants of our method outperform the state-of-the-art. Therefore, enforcing the geometric and spatial constraints is beneficial to the binaural audio generation task. We get the best results when we combine both the losses in our framework.

To further evaluate the utility of our SimBinaural dataset, we next jointly train with both SimBinaural and FAIR-Play,
then test on a challenging split of FAIR-Play 
in which the test scenes are non overlapping, as proposed in \cite{xu2021visually}. 
We compare our method with Augment-PseudoBinaural~\cite{xu2021visually}\footnote{The pre-trained model provided by PseudoBinaural~\cite{xu2021visually} is trained on a different split instead of the standard split from~\cite{gao2019visualsound} and hence it is not directly comparable in Table~\ref{tab:FairPlay-Metric}. We evaluate on the new split in Table~\ref{tab:joint}.} which also uses additional generated training data. Our method with SimBinaural outperforms other methods (Table~\ref{tab:joint}).  This is an important result, as it demonstrates that SimBinaural can be leveraged to improve performance on \emph{real} video.

\vspace*{-0.15in}
\paragraph{User study} 
\begin{wrapfigure}{r}{0.55\textwidth}
    \vspace{-0.2in}
    \begin{tabular}{cc}
        \includegraphics[width=0.525\textwidth]{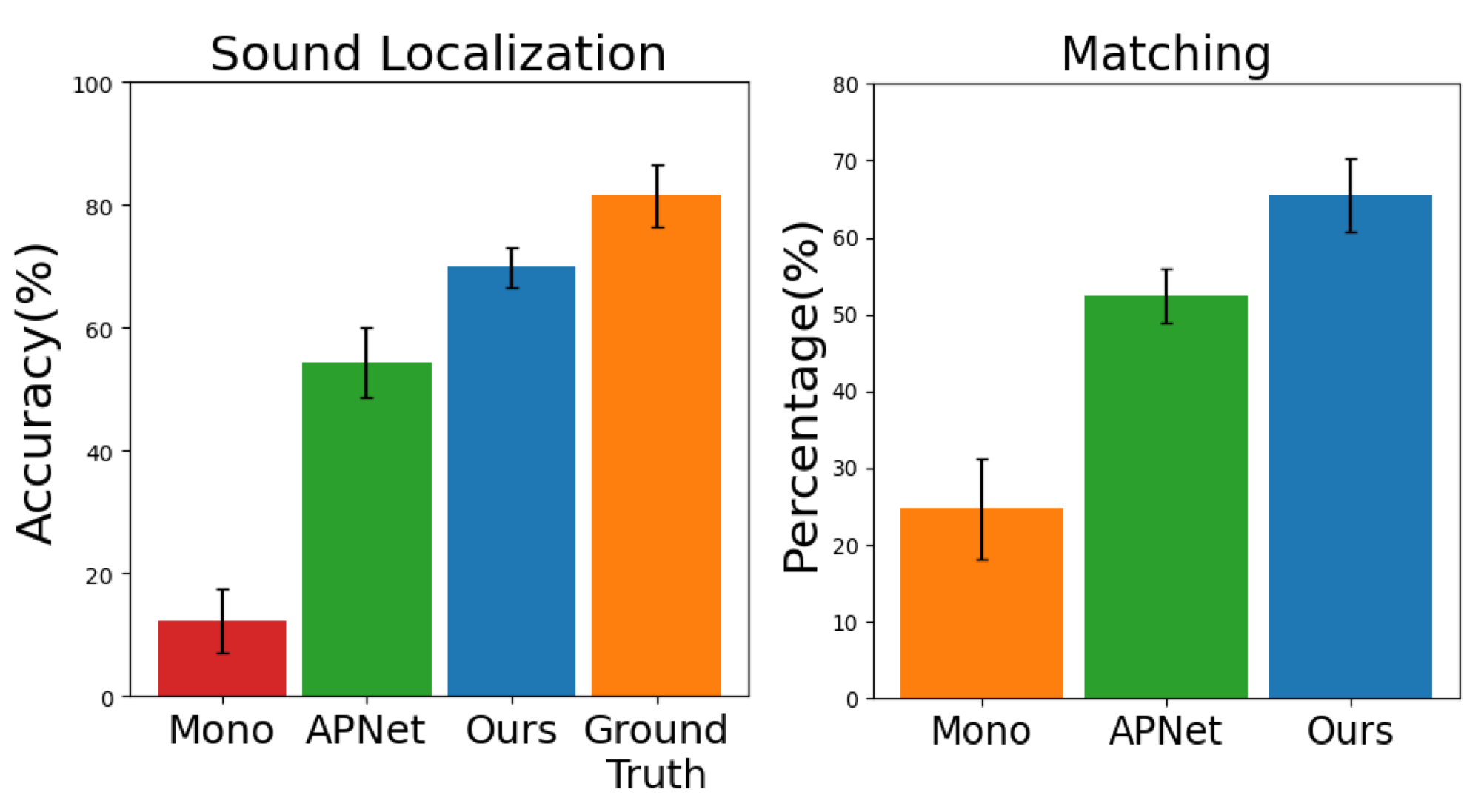}
         \end{tabular}
     \caption{User study results. See text for details. }
     \vspace*{-0.05in}
     \label{fig:user_study}
\end{wrapfigure}
Next, we present two user studies to validate whether the predicted binaural sound does indeed provide an immersive and spatially accurate experience for human listeners. Twenty participants with normal hearing were presented with 20 videos from the test set of the two datasets. They were asked to rate the quality in two ways: 1) users were given only the audio and asked to choose from which direction (left/right/center) they heard the audio; 2) given a pair of audios and a reference frame, the users were asked to choose which audio gives a binaural experience closer to the provided ground truth. As can be seen in Fig.~\ref{fig:user_study}, users preferred our method both for the accuracy of the direction of sound (left) and binaural audio quality (right). 

\begin{figure*}[t]
\centering
\includegraphics[width=0.33\textwidth]{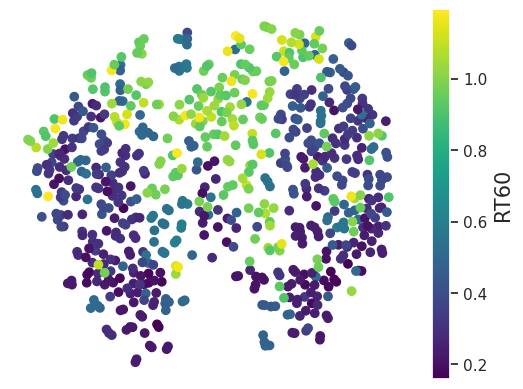}
\includegraphics[width=0.33\textwidth]{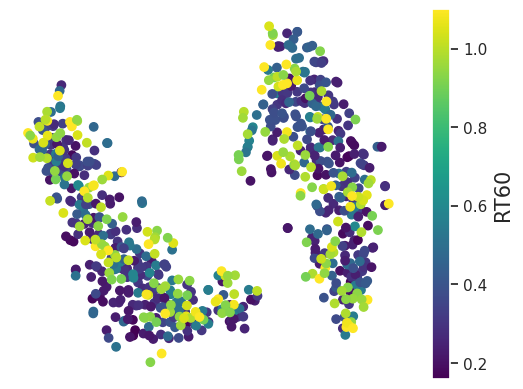}
\includegraphics[width=0.29\textwidth]{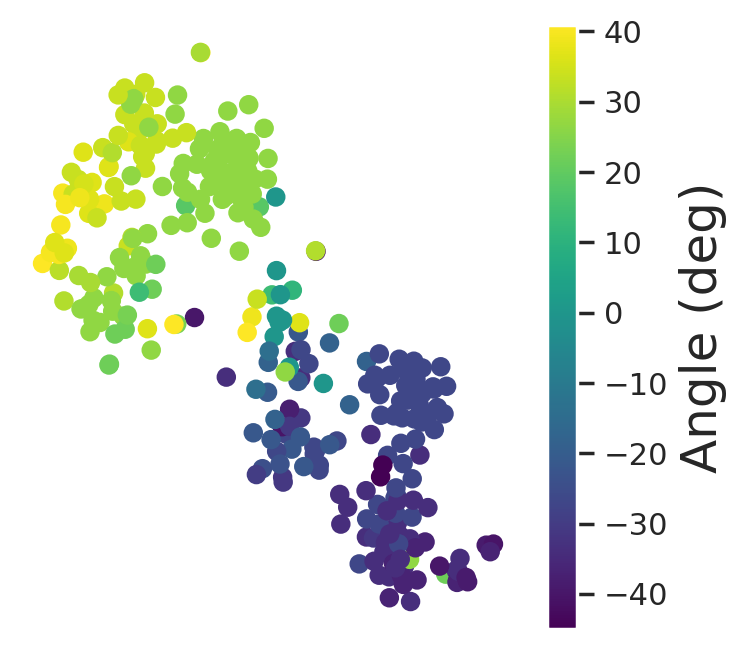}

\vspace*{0.1in}
\caption{t-SNE of visual features colored by $RT_{60}$ for our method (left) and APNet (center); and colored by angle of the object from the center (right). }
\label{fig:tSNE}
\end{figure*}

\begin{figure*}[h!]
\centering
    \includegraphics[width=1\textwidth]{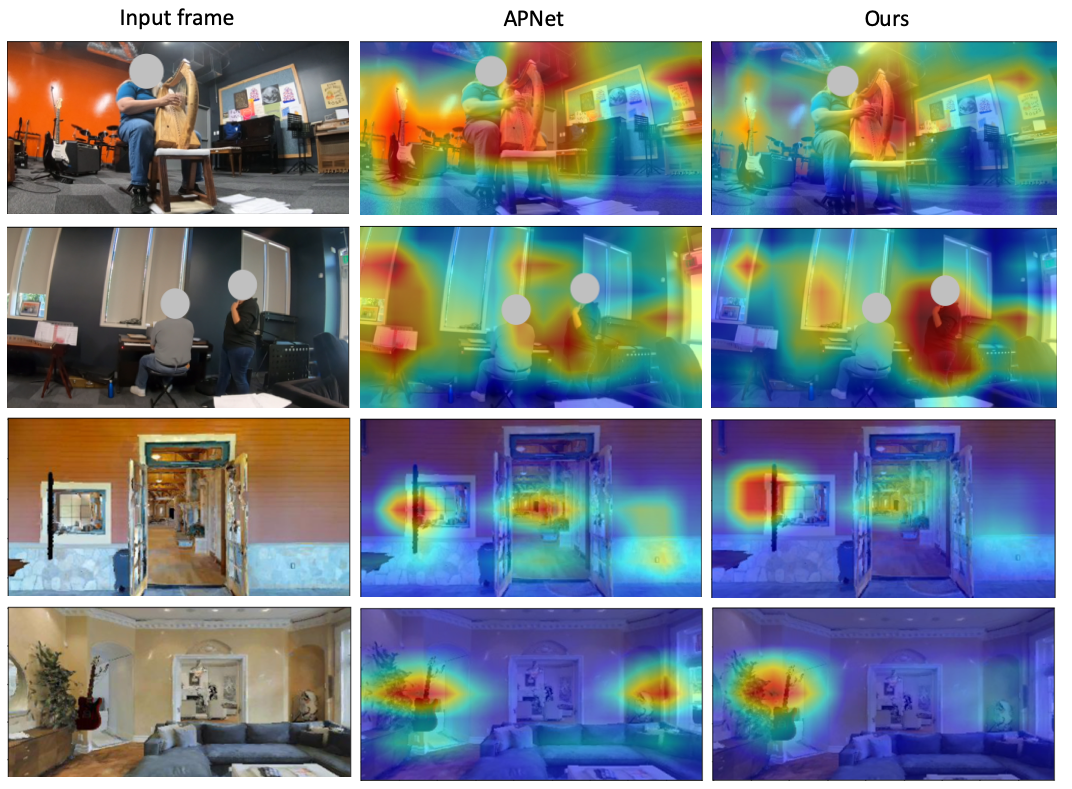}
     \vspace{0.1in}
     \caption{Qualitative visualization of the activation maps for the visual network for APNet~\cite{zhou2020sep} and ours. We can see that while the activation maps for APNet~\cite{zhou2020sep} are diffused and focusing on non-essential parts like objects in the background, our method focuses more on the object/region producing the sound and its location.}
     \label{fig:viz}
\vspace*{0.1in}
\end{figure*}

\vspace*{-0.15in}
\paragraph{Visualization} 
Figure~\ref{fig:tSNE} shows the t-SNE projections~\cite{van2008visualizing} of the visual features from SimBinaural colored by the $RT_{60}$ of the audio clip. While the features from our method (left) can infer the $RT_{60}$ characteristics, the ones from APNet~\cite{zhou2020sep} (center) are randomly distributed. Simultaneously, our features also accurately capture the angle of the object from the center (right). Fig.~\ref{fig:viz} shows the activation maps of the visual network. While APNet produces more diffuse maps, our method localizes the object better within the image. This indicates that the visual features in our method are better at identifying the regions which might be emitting sound to generate more accurate binaural audio.

%===========================================================
\vspace{-0.05in}
\section{Conclusion}
We presented a multi-task approach to learn geometry-aware visual features for mono to binaural audio conversion in videos. Our method exploits the inherent room and object geometry and spatial information encoded in the visual frames to generate rich binaural audio. We also generated a large-scale  video dataset with binaural audio in photo-realistic environments to better understand and learn the relation between visuals and binaural audio. This dataset will be made publicly available to support further research in this direction.  Our state-of-the-art results on two datasets demonstrate the efficacy of our proposed formulation. In future work we plan to explore how semantic models of object categories' sounds could benefit the spatialization task. 

\paragraph{Acknowledgements} Thanks to Changan Chen for help with experiments, Tushar Nagarajan for feedback on paper drafts, and the UT Austin vision group for helpful discussions. UT Austin is supported by NSF CNS 2119115 and a gift from Google. Ruohan Gao was supported by a Google PhD Fellowship.

\bibliography{egbib}

%===========================================================
\newpage
\section*{Appendix}
\vspace*{-0.075in}
\appendix

\section{Supplementary Video}
\vspace*{-0.05in}
In our supplementary video\footnote{\url{http://vision.cs.utexas.edu/projects/geometry-aware-binaural}}, we show (a) examples of our SimBinaural dataset; (b) example results of the binaural audio prediction task on both SimBinaural and FAIR-Play datasets; and (c) examples of the interface for the user studies. 
\vspace*{-0.15in}

\section{RIR Prediction Case Study}
\vspace*{-0.20in}
\begin{figure*}[h!]
\centering
\includegraphics[width=\textwidth]{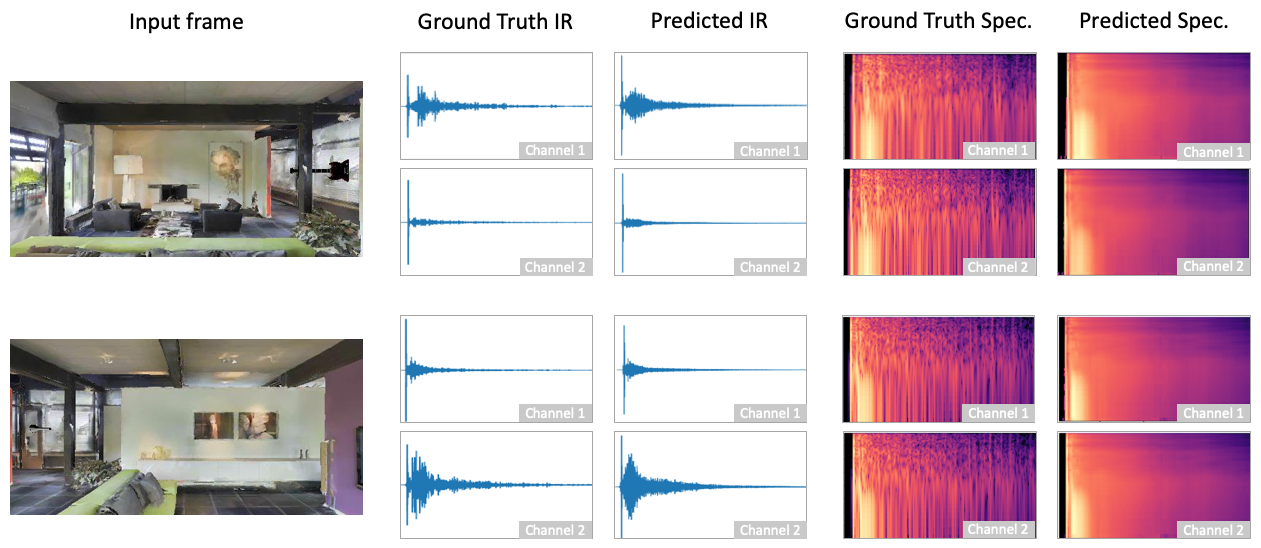}
\vspace*{0.05cm}
\caption{IR Prediction: The first column is the input frame to the encoder. The second column depicts the ground truth IR for the frame and the fourth column is the corresponding spectrogram of this IR. The third and fifth columns show the predicted IR waveform and spectrogram, respectively. This predicted IR waveform is estimated from the spectrogram generated by our network.
}
\vspace*{-0.1in}
\label{fig:ir}
\end{figure*}

We perform a case study on the task of predicting the binaural IR directly from a single visual frame. This helps us evaluate if it is feasible to learn this information just from a visual frame, so that it can be then used for our task as in Sec. 3.2 of the main paper. We predict the acoustic properties of the room by looking at one snapshot of the scene. We predict the magnitude spectrogram of the IR for the two channels. We also obtain the predicted waveform of the IR using the Griffin-Lim algorithm \cite{griffin1984signal}. Figure~\ref{fig:ir} shows qualitative examples of predictions from the test set. It can be seen that we can get a fairly accurate general idea of the IR, and the difference between the response in each channel is also captured.

To evaluate if we capture the materials and geometry effectively, we train another task to predict the reverberation time $RT_{60}$ of the IR from the visual frame. A more accurate prediction of $RT_{60}$ means that our network understands how the wave will interact with the room and materials and whether it takes more or less time to decay. We formulate this as a classification task and discretize the range of the $RT_{60}$ into 10 classes, each with roughly equal number of samples. We then use a classifier to predict this range class of $RT_{60}$ using only the visual frame as input. The classifier, consisting of a ResNet-18, has a test accuracy of 61.5\% which demonstrates the networks' ability to estimate the $RT_{60}$ range quite well.
\vspace*{-0.1in}

\section{SimBinaural dataset details}
\vspace*{-0.05in}
To construct the dataset, we insert diverse 3D models of various instruments like guitar, violin, flute etc. and other sound-making objects like phones and clocks into the scene. Each object has multiple models of that class for diversity, so we do not associate a sound with a particular 3D model. We have a total of 35 objects from 11 classes.

To generate realistic binaural sound in the environment as if it is coming from the source location and heard at the camera position, we convolve the appropriate SoundSpaces~\cite{chen2020soundspaces} room impulse response with an anechoic audio waveform (e.g., a guitar playing for an inserted guitar 3D object). We use sounds recorded in anechoic environments, so there is no existing reverberations to affect the data. The sounds are obtained from Freesound \cite{font2013freesound} and OpenAIR data \cite{murphy2010openair} to form a set of 127 different sound clips spanning the 11 distinct object categories. Using this setup, we capture videos with simulated binaural sound.  

The virtual camera and attached microphones are moved along trajectories such that the object remains in view, leading to diversity in views of the object and locations within each video clip. Using ray tracing, we ensure that the object is in view of the camera, and the source positions are densely sampled from the 3D environments. For a particular video, we use a fixed source position and the agent traverses a random path. The view of the object changes throughout the video as the camera moves and rotates, so we get diverse orientations of the object and positions within a video frame, for each video. The camera moves to a new position every 5 seconds and has a small translational motion during the five-second interval. The videos are generated at 5 frames per second, the average length of the videos in the dataset is 30.3s and the median length is 20s. 
\vspace*{-0.1in}

\section{Implementation Details}
All networks are written in PyTorch \cite{NEURIPS2019_9015}. The backbone network is based upon the networks used for 2.5D visual sound~\cite{gao2019visualsound} and APNet~\cite{zhou2020sep}. The visual network is a ResNet-18~\cite{he2016deep} with the pooling and fully connected layers removed. The U-Net consists of 5 convolution layers for downsampling and 5 upconvolution layers in the upsampling network and include skip connections. The encoder for spatial coherence follows the same architecture as the U-Net encoder for the audio feature extraction.  The classifier combines the audio and visual features and uses a fully connected layer for prediction. The generator network is adapted from GANSynth \cite{engel2019gansynth}, modified to fit the required dimensions of the audio spectrogram.  

To preprocess both datasets, we follow the standard steps from~\cite{gao2019visualsound}. We resampled all the audio to 16kHz and computed the STFT using a FFT size of 512, window size of 400, and hop length of 160. For training the backbone, we use 0.63s clips of the 10s audio and the corresponding frame. Frames are extracted at 10fps. The visual frames are randomly cropped to 448 $\times$ 224. For testing, we use a sliding window of 0.1s to compute the binaural audio for all methods. 

We use the Adam optimizer \cite{adamsolver} and a batch size of 64. The initial learning rates are 0.001 for the audio and fusion networks, and 0.0001 for all other networks. We trained the FAIR-Play dataset for 1000 epochs and SimBinaural for 100 epochs. We train the RIR prediction separately and use the weights for initialization while training jointly. The $\delta$ for choice of frame is set to 1s and the $\lambda$'s used are set based on validation set performance to $\lambda_B=10, \lambda_S=1, \lambda_G=0.01, \lambda_P=1$.  

\section{Additional Ablations}
Table 2 in the main paper illustrates that adding each component of our method individually to the visual features helps improve the binaural audio quality performance. Table~\ref{tab:addition} provides additional analysis to evaluate the combination of different constraints with the backbone for the SimBinaural Position-Split. The constraints complement each other to learn better visual features, leading to better audio performance. 

\begin{table}
\centering
\begin{tabular}{ccc}\small
Method  & STFT   & ENV    \\ \hline
Spatial+Geometric & 0.724  & 0.118  \\
IR Pred+Geometric & 0.707  & 0.117  \\
IR Pred+Spatial   & 0.702  & 0.117
\end{tabular}
\vspace*{0.1in}
\caption{Results on SimBinaural Position-Split with different combinations of constraints.
}
\vspace{-0.1in}
\label{tab:addition}
\end{table}

\end{document}